%% file: ms.tex
\documentclass{article}

\usepackage{arxiv_preprint}
\usepackage[numbers, sort&compress]{natbib}

\input{packages}
\input{macros}

\title{Bayesian Estimation of Differential Privacy}

\author{
  Santiago Zanella-B{\'e}guelin\thanks{Joint main author.} \\
  Microsoft Research \\
  Cambridge, UK \\
  \texttt{santiago@microsoft.com} \\
  \And
  Lukas Wutschitz\footnotemark[1] \\
  Microsoft \\
  Cambridge, UK \\
  \texttt{luwutsch@microsoft.com} \\
  \And
  Shruti Tople\footnotemark[1] \\
  Microsoft Research \\
  Cambridge, UK \\
  \texttt{shtople@microsoft.com} \\
  \And
  Ahmed Salem \\
  Microsoft Research \\
  Cambridge, UK \\
  \texttt{t-salemahmed@microsoft.com} \\
  \And 
  Victor R{\"u}hle \\
  Microsoft \\
  Cambridge, UK \\
  \texttt{virueh@microsoft.com} \\
  \And
  Andrew Paverd \\
  Microsoft Research\\
  Cambridge, UK \\
  \texttt{anpaverd@microsoft.com} \\
  \And
  Mohammad Naseri \\
  University College London \\
  London, UK \\
  \texttt{mohammad.naseri.19@ucl.ac.uk} \\
  \And
  Boris K{\"o}pf \\
  Microsoft Research \\
  Cambridge, UK \\
  \texttt{bokoepf@microsoft.com}
  \And
  Daniel Jones \\
  Microsoft \\
  Cambridge, UK \\
  \texttt{jonesdaniel@microsoft.com}
}

\begin{document}

\maketitle

\begin{abstract}
Algorithms such as Differentially Private SGD enable training machine learning models with formal privacy guarantees. However, there is a discrepancy between the protection that such algorithms guarantee in theory and the protection they afford in practice.
An emerging strand of work empirically estimates the protection afforded by differentially private training as a confidence interval for the privacy budget $\epsilon$ spent on training a model.
Existing approaches derive confidence intervals for $\epsilon$ from confidence intervals for the false positive and false negative rates of membership inference attacks. Unfortunately, obtaining narrow high-confidence intervals for $\epsilon$ using this method requires an impractically large sample size and training as many models as samples.
We propose a novel Bayesian method that greatly reduces sample size, and adapt and validate a heuristic to draw more than one sample per trained model.
Our Bayesian method exploits the hypothesis testing interpretation of differential privacy to obtain a posterior for $\epsilon$ (not just a confidence interval) from the joint posterior of the false positive and false negative rates of membership inference attacks. For the same sample size and confidence, we derive confidence intervals for $\epsilon$ around 40\% narrower than prior work.
The heuristic, which we adapt from label-only DP, can be used to further reduce the number of trained models needed to get enough samples by up to 2 orders of magnitude.
\end{abstract}

\newpage

\input{introduction}

\input{background}

\input{approach}

\input{parameter_eval.tex}

\input{heuristics_eval}

\input{related}

\input{conclusion}


\input{ms.bbl}
\newpage
\appendix

\input{intervals}

\input{mia}

\input{density}

\input{tables}

\clearpage
\input{convergence}

\end{document}

%% file: packages.tex
\usepackage[utf8]{inputenc} 
\usepackage[T1]{fontenc}    
\usepackage[pagebackref]{hyperref}
\usepackage{hypernat}
\usepackage{url}            
\usepackage{booktabs}       
\usepackage{amsfonts}       
\usepackage{amsthm}
\usepackage{amsmath}
\usepackage{array}
\usepackage{mathtools}
\usepackage{bm}
\usepackage{xspace}
\usepackage{nicefrac}       
\usepackage{microtype}      
\usepackage{xcolor,colortbl}
\usepackage{paralist}
\usepackage{multirow}
\usepackage{graphicx}
\usepackage{siunitx}
\usepackage{pdfpages}
\usepackage{adjustbox}
\usepackage[algo2e]{algorithm2e}
\usepackage{subfig}
\usepackage{tikz}
\usepackage{wrapfig}
\usepackage{doi}
\usepackage{lscape}
\usepackage{cleveref}

\crefname{table}{Table}{Tables}

\usepackage[disable]{todonotes}
\usepackage[compact]{titlesec}
\titlespacing{\section}{0pt}{*0}{*0}
\titlespacing{\subsection}{0pt}{*0}{*0}
\titlespacing{\subsubsection}{0pt}{*0}{*0}
\tikzset{>=latex} 
\usepackage{pgfplots} 
\usepackage[outline]{contour} 
\contourlength{1.2pt}
\usetikzlibrary{positioning,calc}
\usetikzlibrary{backgrounds}

\colorlet{mydarkblue}{blue!30!black}

\usepgfplotslibrary{fillbetween}
\usetikzlibrary{patterns}
\pgfplotsset{compat=1.12} 

\def\N{50}

%% file: macros.tex
\theoremstyle{plain}
\newtheorem{theorem}{Theorem}[section]

\theoremstyle{definition}
\newtheorem{definition}[theorem]{Definition}

\theoremstyle{remark}

\definecolor{Blue3}{HTML}{0000CD}
\definecolor{Green4}{HTML}{008B00}
\definecolor{Red3}{HTML}{CD0000}
\definecolor{orange}{rgb}{0.8, 0.47, 0.196}
\definecolor{Red}{rgb}{1, 0, 0}

\definecolor{Blue3}{HTML}{0000CD}
\definecolor{Green4}{HTML}{008B00}
\definecolor{Red3}{HTML}{CD0000}
\definecolor{orange}{rgb}{0.8, 0.47, 0.196}
\definecolor{Red}{rgb}{1, 0, 0}

\RestyleAlgo{ruled}
\newenvironment{experiment}[1][htb]
  {
   \begin{algorithm2e}[#1]%
  }{\end{algorithm2e}}

\renewcommand{\Pr}[1]{\mathrm{Pr}\left[ #1 \right]}
\newcommand{\A}{\mathcal{A}}

\newcommand{\guess}[1]{\tilde{#1}}
\newcommand{\Adv}[1]{\mathrm{Adv}^{\mathrm{#1}}}

\newcommand{\iid}{i.i.d.\@\xspace}
\newcommand{\eg}{e.g.\@\xspace}
\newcommand{\ie}{i.e.\@\xspace}
\newcommand{\wrt}{w.r.t.\@\xspace}
\newcommand{\lb}[1]{\ensuremath{#1_{-}}}
\newcommand{\ub}[1]{\ensuremath{#1_{+}}}

\newcommand{\FP}{\textnormal{FP}\xspace}
\newcommand{\FN}{\textnormal{FN}\xspace}
\newcommand{\TP}{\textnormal{TP}\xspace}
\newcommand{\TN}{\textnormal{TN}\xspace}
\newcommand{\FPR}{\textnormal{FPR}\xspace}
\newcommand{\FNR}{\textnormal{FNR}\xspace}
\newcommand{\FNRFPR}{(\FNR, \FPR)\xspace}
\newcommand{\eeps}{\textnormal{e}^{\varepsilon}}
\newcommand{\empepsilon}{\ensuremath{\hat{\epsilon}}\xspace}
\newcommand{\dd}{\textnormal{d}}
\newcommand{\lo}{\textnormal{LO}}
\newcommand{\hi}{\textnormal{HI}}
\newcommand{\Bin}{\mathrm{Bin}}
\newcommand{\Beta}{\mathrm{Beta}}
\newcommand{\BetaPPF}{\mathrm{B}}

\renewcommand{\epsilon}{\varepsilon}

\newcommand{\eqdef}{\coloneqq}

\newcolumntype{P}[1]{>{\centering\arraybackslash}p{#1}}
\newcolumntype{M}[1]{>{\centering\arraybackslash}m{#1}}

\renewcommand*{\backref}[1]{}
\renewcommand*{\backrefalt}[4]{%
  \ifcase #1%
    \or Cited on p.~#2.%
  \else Cited on pp.~#2%
  \fi%
}

%% file: introduction.tex
\section{Introduction}
\label{sec:introduction}

The use of machine learning in industries such as healthcare and finance requires strong and auditable safeguards against leakage of sensitive training data.
Differentially Private (DP) training using algorithms such as DP-SGD~\cite{Abadi:2016} and PATE~\cite{Papernot:2017} partially addresses this concern by bounding the amount of information that can be leaked through models.
However, there is a gap between the degree of protection that DP training offers {\em in theory}, and the protection it offers {\em in practice}.
For example, DP training with a privacy budget of $\epsilon \approx 4$, a common choice in practice~\cite{Desfontaines:2021}, cannot rule out membership inference attacks~\cite{Humphries:2021}. 
Nonetheless, DP training with such large budgets can effectively defeat attacks in many practical scenarios~\cite{Song:2019,Carlini:2019,Jayaraman:2019,Zanella-Beguelin:2020}.
The reason for this discrepancy is that provable DP bounds~\cite{Gopi:2021} hold up to extremely powerful adversary models (\eg, in the case of DP-SGD, adversaries that can see and tamper with intermediate gradients), and so overestimate the privacy risks of weaker adversaries that matter in practice. 

Without any information beyond provable DP bounds, practitioners must either err on the side of caution and use unnecessarily small privacy budgets which hurt utility, or take the risk of using larger budgets based on a guess of the privacy they provide.
To resolve this conflict, an emerging strand of work aims to measure the protection afforded by DP training against specific adversaries by computing statistical estimates \empepsilon for the privacy budget spent~\cite{Hyland:2019,Jagielski:2020,Carlini:2021b,Malek:2021}.
A confidence interval for the privacy budget $\empepsilon$ spent by a training pipeline can be calculated from estimates of the false positive and false negative rates of membership inference attacks ran against models trained with it. 
However, existing approaches exhibit two limitations that prevent them from scaling to large models, or to large numbers of models, as required for architecture search and hyperparameter tuning:
\begin{asparaenum}
\item On the statistical side, current approaches bound the false positive and false negative rates separately using Clopper-Pearson (CP) confidence intervals, which notoriously underestimate coverage and require a large sample size to draw conclusions with high confidence. In fact, for sample sizes considered in prior work, confidence intervals for $\empepsilon$ derived from CP intervals are so wide that they often include 0 and the provable upper bound for DP models~\citep[Fig.~1]{Carlini:2021b}.\label{it:statistics}

\item On the computational side, current approaches require that {\em each sample} be obtained from a model that is independently trained. An exception is~\cite{Malek:2021}, which proposes a heuristic for estimating \emph{label-only} differential privacy that draws $m > 1$ samples from a {\em single} model.\label{it:heuristics}
\end{asparaenum}

To overcome the first limitation, we propose a novel Bayesian approach that is more precise and thus requires fewer samples to converge to meaningful estimates.
In line with prior art~\cite{Jagielski:2020,Carlini:2021b}, we derive estimates of \empepsilon based on estimates of the false positive and false negative rates of membership inference attacks.
Unlike previous approaches which derive estimates from {\em separate} confidence intervals for each quantity, we model their \emph{joint distribution}. 
Exploiting the hypothesis testing interpretation of differential privacy, we use this joint distribution to compute a posterior distribution for \empepsilon, from which we derive significantly tighter credible intervals.

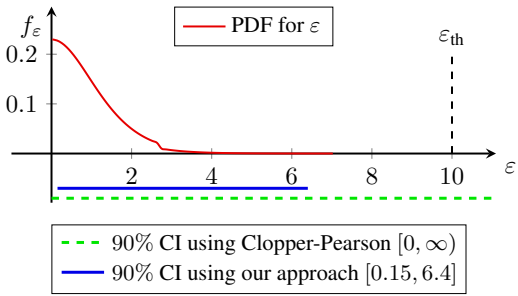
\begin{wrapfigure}[19]{r}{0.52\textwidth}
	\centering
	\scalebox{0.94}{
	\input{tikz/empirical_vs_theoretical.tex}
	}
	\caption{
		Comparison of the posterior PDF $f_\varepsilon$ using our Bayesian approach and the upper bound $\varepsilon_{\textnormal{th}}$ obtained from a state-of-the-art DP accountant \cite{Gopi:2021} for a CNN trained on CIFAR10 with $\delta=10^{-5}$.
		The empirical estimation suggests stronger privacy than the theoretical guarantee.
		At the bottom, we illustrate the reduction in uncertainty of the $90\%$ credible interval using our Bayesian approach over the Clopper-Pearson interval.
	}
	\label{fig:illustration}
\end{wrapfigure}

We evaluate the performance of this Bayesian approach in numeric simulations and in experiments on text and vision classifiers. For both settings, we compare equal-tailed credible intervals for \empepsilon obtained using our approach with confidence intervals for \empepsilon derived from Clopper-Pearson and Jeffreys intervals for false positive and false negative rates.
In our experiments we observe a reduction in interval width of up to \textbf{40\%} with respect to prior work for the same number of samples. Figure~\ref{fig:illustration} illustrates these gains.
Our approach enables us to draw conclusions that are as significant as prior work but with significantly fewer samples.

To overcome the second limitation, we adapt the heuristic of \citet{Malek:2021} to \emph{full} differential privacy.
We use our Bayesian approach to compare estimates computed using the heuristic for different values of $m$ to the baseline $m=1$.
Specifically, we investigate whether (and under what circumstances) this heuristic provides faithful estimates, to identify a suitable trade-off between $N$, the total number of samples used for the estimate, and $N/m$, the number of independent models that need to be trained. 
To this end, we run experiments on text and vision classifiers for $m=10,100, 1000$, and we compare the posterior distributions of $\empepsilon$ with that corresponding to the baseline $m=1$.
Our results show that for training pipelines that satisfy differential privacy, the heuristic can reduce the number of models required to be trained by up to \textbf{2 orders of magnitude} while still yielding faithful estimates.

\paragraph{Summary of contributions}
We propose a novel Bayesian approach that yields high-confidence estimates of the differential privacy budget spent by training pipelines.
We show through experiments on text and vision classifiers that this approach translates into privacy estimates that are significantly tighter than using existing approaches.
Furthermore, we validate a previously proposed heuristic that can provide an $m$-fold reduction in the number of models that need to be trained.
Combined, our Bayesian approach and this heuristic can significantly reduce the computational cost of obtaining meaningful privacy estimates.

%% file: tikz/empirical_vs_theoretical.tex
\begin{tikzpicture}[inner frame sep=0]
  \def\xmax{10};
  \def\ymin{{(-0.099)}};
  \def\ymax{{(0.26)}};
  \def\epsth{{(10.0)}};
  
  \begin{axis}[every axis plot post/.append style={
               mark=none,domain={-0.05*(\xmax)}:{1.08*\xmax},samples=\N,smooth},
               xmin={-0.1*(\xmax)}, xmax=\xmax,
               ymin=\ymin, ymax=\ymax,
               axis lines=middle,
               axis line style=thick,
               enlargelimits=upper, 
               xlabel=$\varepsilon$,
               ylabel={$f_{\varepsilon}$},
               every axis x label/.style={at={(current axis.right of origin)},anchor=north west},
               every axis y label/.style={at={(current axis.above origin)},anchor=north east,align=center},
               y=200pt,
               legend style={at={(0.5, 1.0)},anchor=north,legend cell align=left} %
              ]
    
    \addplot[thick,black!10!red] table[x=epsilon, y=pdf, col sep=tab] {data/epsilon_pdf.tsv};
    \addlegendentry{\small PDF for $\varepsilon$};
    \addplot[very thick, black!10!blue] coordinates {(0.15, -0.07) (6.4, -0.07)};
    \label{g:CIours}
    \addplot[very thick, black!10!green, dashed] coordinates {(0.0, -0.09) (11.0, -0.09)};
    \label{g:CICP}

    \addplot[black,dashed,thick]
      coordinates {(\epsth, 0.0) (\epsth, 0.2)}
      node[above=0pt,pos=1] {$\varepsilon_{\textnormal{th}}$};
   
  \end{axis}
  \node [draw,fill=white] at (3.5,-0.8) {\shortstack[l]{
    \ref*{g:CICP} \small  $90\%$ CI using Clopper-Pearson $[0,\infty)$ \\
    \ref*{g:CIours} \small $90\%$ CI using our approach $[0.15, 6.4]$}};
\end{tikzpicture}

%% file: background.tex
\section{Preliminaries}
\label{sec:background}

In this section we introduce the notation used throughout the paper, recall the definition of $(\epsilon,\delta)$-differential privacy and its hypothesis testing interpretation, then overview membership inference attacks and their relation to differential privacy.

\subsection{Notation}

We use calligraphy font for randomized algorithms (\eg, $\mathcal{T}$) and distributions (\eg, $\mathcal{D}$), and uppercase serif font for lists and sets (\eg, $S$). We use $z \sim \mathcal{D}$ to denote a sample $z$ drawn from $\mathcal{D}$ and $S \sim \mathcal{D}^n$ to denote a list $S$ of $n$ samples independently drawn from $\mathcal{D}$. $b \sim \{0,1\}$ denotes a fair coin sample, \ie, a bit $b$ sampled uniformly from $\{0,1\}$.
Adversary algorithms (\eg, $\mathcal{A}_1, \mathcal{A}_2$) are randomized procedures that share mutable state, although for clarity we often include redundant arguments.
We formalize probabilistic experiments as sequential pseudocode and write $\Pr{\mathrm{Exp}(\cdots) : A}$ for the probability of event $A$ in experiment $\mathrm{Exp}$.
Table~\ref{tab:notation} summarizes this notation.

\begin{table}[ht]
  \centering
  \begin{tabular}{@{}l@{~~}l@{}}
  \toprule
  \bf Notation    & \bf Description \\
  \midrule
  $\mathcal{T}$   & A stochastic training algorithm \\
  $\mathcal{D}$   & Distribution over samples \\
  $\mathcal{D}^n$ & Distribution of $n$ independent samples from $\mathcal{D}$ \\
  $\A$, $\A_1$, $\A_2$   & Adversary procedures sharing mutable state \\
  $z \sim \mathcal{D}$   & Draw a sample $z$ from $\mathcal{D}$ \\
  $S \sim \mathcal{D}^n$ & Draw a list $S$ of $n$ independent samples from $\mathcal{D}$ \\
  $b \sim \{0,1\}$       & Sample a bit $b$ uniformly \\
  $y \gets \mathcal{P}(\vec{x})$ & Call $\mathcal{P}$ with arguments $\vec{x}$ and assign result to $y$ \\
  \bottomrule
  \end{tabular}
  \caption{Summary of notation}
  \label{tab:notation}
\end{table}

\subsection{Approximate Differential Privacy}

\begin{definition}[Approximate Differential Privacy]
Let $\epsilon > 0$ and $\delta \in [0,1]$. A mechanism $\mathcal{T} : X \to Y$ is $(\epsilon,\delta)$-differentially private with respect to an adjacency relation $R \subseteq X \times X$ if for any $(x, x') \in R$ and any $O \subseteq Y$,
\begin{equation*}
  \Pr{\mathcal{T}(x) \in O} \leq e^\epsilon\ \Pr{\mathcal{T}(x') \in O} + \delta \ .
\end{equation*}
\end{definition}
The mechanisms we study are machine learning training algorithms of the form $\mathcal{T} : X^n \to \Theta$ that produce model weights $\theta \in \Theta$ given a dataset $S$ of $n$ examples from $X$. 
We refer to $S$ as the training dataset of $\theta$, which under normal circumstances is composed of \iid examples drawn from some underlying distribution $\mathcal{D}$ with support $X$.
We consider two training datasets as adjacent if one can be obtained from the other by substituting a single element. 
This corresponds to \emph{bounded differential privacy}~\citep{Kifer:2011}.

\pagebreak

\subsection{Hypothesis Testing Characterization of Differential Privacy}

\begin{wrapfigure}[20]{r}{0.52\textwidth} 
  \centering
  \vspace{-15pt}
  \includegraphics[width=0.52\textwidth]{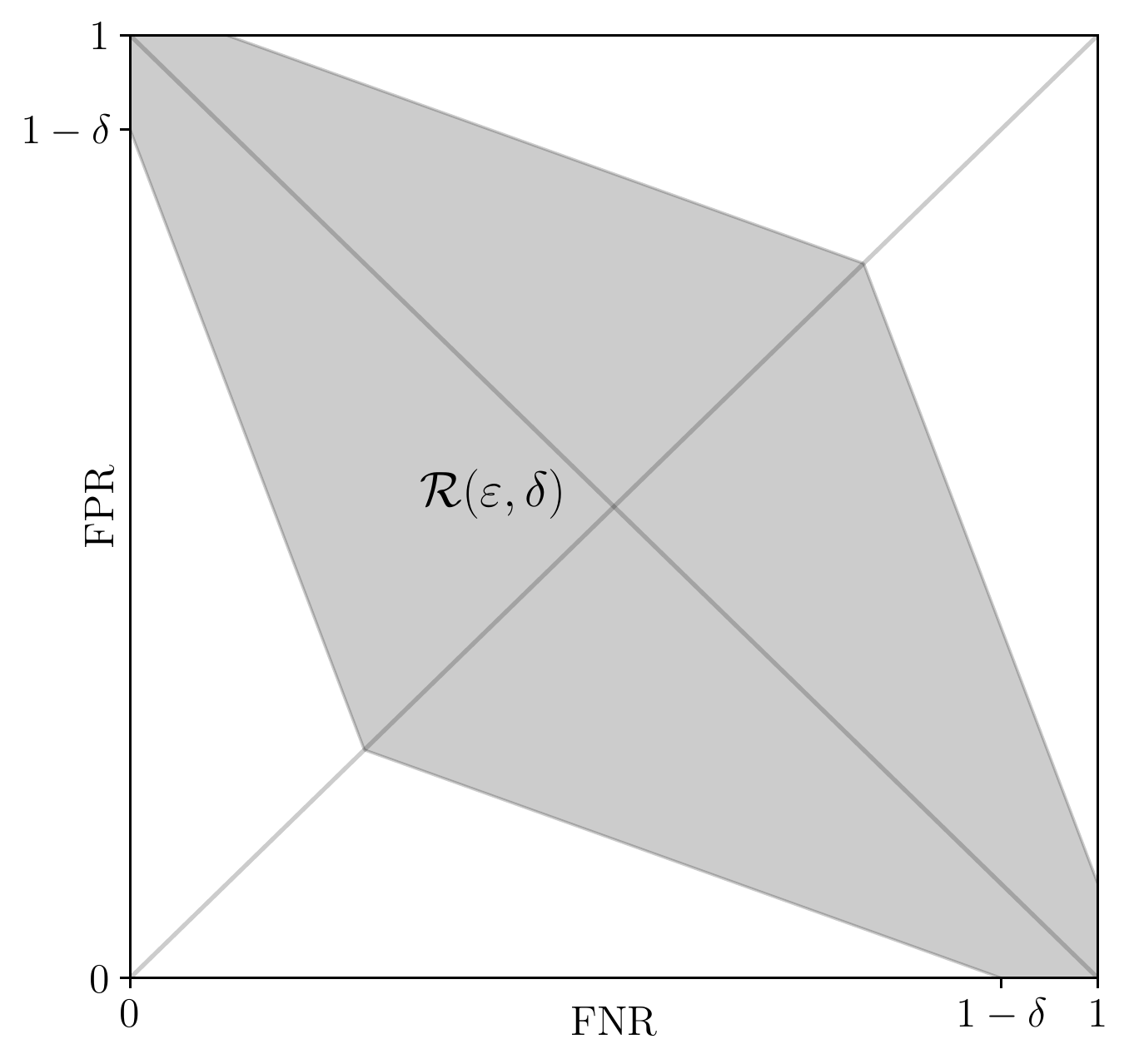}
  \caption{Privacy region $\mathcal{R}(\epsilon,\delta)$. The region grows with $\varepsilon$ and covers the unit square as $\varepsilon$ tends towards $\infty$.
  }
  \label{fig:privacy_region}
\end{wrapfigure}
Consider a run of a mechanism $\mathcal{T} : X \to Y$ that outputs some $y \in Y$ when given one of two adjacent inputs $D, D'$.
We can recast the differential privacy of $\mathcal{T}$ as a hypothesis test where the null hypothesis is that the input was $D$ and the alternative hypothesis is that it was $D'$.
The test rejects the null hypothesis when $y$ is in a rejection region $R$.
A Type-I error (false positive) occurs when the null hypothesis is true but is rejected, with probability $\Pr{\mathcal{T}(D) \in R}$.
A Type-II error (false negative) occurs when the null hypothesis is false but is not rejected, with probability $\Pr{\mathcal{T}(D') \in \overline{R}}$.

The following theorem from~\cite{Kairouz:2017} characterizes $(\epsilon,\delta)$-differentially privacy in terms of conditions on the false positive and false negative rates of hypothesis tests.
This extends an earlier result from \cite{Hall:2013} that only shows that the conditions are necessary.

\begin{theorem}\label{thm:hyptest}
A mechanism $\mathcal{T} : X \to Y$ is $(\epsilon,\delta)$-differentially private if and only if for all adjacent inputs $D,D'$ and all $R \subseteq Y$, the following conditions are met
\begin{align*}
  \Pr{\mathcal{T}(D) \in R} + \textnormal{e}^\epsilon\,\Pr{\mathcal{T}(D') \in \overline{R}} \geq 1 - \delta \; , \\
  \Pr{\mathcal{T}(D') \in \overline{R}} + \textnormal{e}^\epsilon\,\Pr{\mathcal{T}(D) \in R} \geq 1 - \delta \; .
\end{align*}
\end{theorem}

A distinguisher that observes the output of an $(\epsilon,\delta)$-differentially private mechanism $\mathcal{T}$ and makes a guess as to which hypothesis is true implicitly defines a rejection region. The set of false positive and false negative rates achievable by distinguishers, or equivalently, the set of Type-I and Type-II errors for any rejection region must be included in the \emph{privacy region} $\mathcal{R}(\epsilon,\delta)$, defined as follows:
\begin{equation*}
\mathcal{R}(\epsilon,\delta) \!\eqdef\! \{ (x,y) \mid
  x + \eeps y \geq 1 - \delta     \land y + \eeps x \geq 1 - \delta \land
  y + \eeps x \leq \eeps + \delta \land x + \eeps y \leq \eeps + \delta
\} \; .
\end{equation*}

Figure~\ref{fig:privacy_region} illustrates the privacy region $\mathcal{R}(\epsilon,\delta)$.
It is symmetric \wrt the $\FNR = 1 - \FPR$ line because if a rejection region $Y$ achieves $(\FNR,\FPR)$, its complement $\overline{Y}$ achieves $(1-\FNR,1-\FPR)$.
It is also symmetric \wrt the $\FNR = \FPR$ line because the adjacency relation is symmetric and so positive and negative instances are interchangeable.

\subsection{Differential Privacy Estimates from Membership Inference}

\begin{wrapfigure}[9]{r}{0.47\textwidth}
\vspace{-10pt}
\begin{minipage}{0.47\textwidth}
  \begin{experiment}[H]
    \DontPrintSemicolon
    \caption[F]{$\textrm{IND-MIA}$}
    \KwIn{$\mathcal{T}, \mathcal{D}, n, \A$}
      $S, z_0, z_1 \gets \A_1(\mathcal{T}, \mathcal{D}, n)$
      \tcp*[r]{$|S| = n - 1$}
      $b \sim \{0,1\}$\;
      $\theta \gets \mathcal{T}(S \cup \{z_b\})$\;
      $\guess{b} \gets \A_2(\mathcal{T}, \mathcal{D}, n, \theta, S, z_0, z_1)$
  \label{exp:ind-mia}
  \end{experiment}
\end{minipage}
\end{wrapfigure}

Membership inference attacks (MIA) try to determine whether samples belong to the training dataset of a model. 
Rather than the standard MIA experiment from the literature (see \eg~\cite{Yeom:2018}), we consider more powerful DP distinguishers as in Experiment~\ref{exp:ind-mia}, which can select a base training dataset $S$ and challenge points $z_0,z_1$.

A MIA such as Experiment~\ref{exp:ind-mia} defines a hypothesis test. Its false negative and false positive rates can be written as follows:
\begin{eqnarray*}
\FNR \eqdef \Pr{\mathrm{IND-MIA} : \guess{b} = 0 \mid b = 1} \; , \qquad 
\FPR \eqdef \Pr{\mathrm{IND-MIA} : \guess{b} = 1 \mid b = 0} \\
\end{eqnarray*}

We use this interpretation to bound the empirical privacy parameter \empepsilon of a training algorithm for a fixed $\delta$.
The idea is that any false positive and false negative rates \FNRFPR serves as a counterexample for the training pipeline being $(\epsilon, \delta)$-differentially private for every $\epsilon$ such that $\FNRFPR \not\in \mathcal{R}(\epsilon, \delta)$. So, a lower bound for \empepsilon is given by   
\begin{equation*}
\lb{\empepsilon} = \sup \{ \epsilon \in \mathbb{R}^+ \mid (\FNR,\FPR) \not\in \mathcal{R}(\epsilon,\delta)\}
\end{equation*}
Assuming $\FPR, \FNR \neq 0$ and $\FPR, \FNR \leq 1 - \delta$, this is
\begin{equation}
\lb{\empepsilon} = \max \left\{ \log \frac{1 - \delta - \FPR}{\FNR},
                             \log \frac{1 - \delta - \FNR}{\FPR} \right\}
\label{eqn:lowerbound}
\end{equation}
Previous work~\citep{Jagielski:2020,Carlini:2021} uses a Monte Carlo approach to estimate \FPR and \FNR with Clopper-Pearson confidence intervals and derives estimates for \empepsilon based on that. See Appendix~\ref{sec:montecarlo} for details.

%% file: approach.tex
\section{A Bayesian Approach to Privacy Estimates}
\label{sec:approach}

In this section we present a novel Bayesian approach to privacy estimates that models false positive and false negative rates as independent binomial proportions with non-informative Jeffreys priors.
We first present Jeffreys intervals, derived from the same model, as an alternative to Clopper-Pearson intervals. We then present a much more precise method that directly computes credible intervals from the posterior distribution of \FNRFPR.

\subsection{Jeffreys Intervals}

Jeffreys intervals have roots in Bayesian analysis, achieve good probability matching properties, and are particularly recommended as one-sided intervals~\citep[p.68]{Cai:2005}.
Their Bayesian derivation uses a non-informative conjugate prior for the binomial proportion $p$, resulting in the model
\begin{equation}
\begin{aligned}
        p &\sim \Beta(1/2, 1/2) \\
    k | p &\sim \Bin(N, p) \\
    p | k &\sim \Beta(1/2 + k, 1/2 + N - k)
\end{aligned}
\end{equation}

The upper-limit of the one-sided $100(1-\alpha)\%$ Jeffreys interval is the $1-\alpha$ quantile of the posterior $p | k$, that is $\BetaPPF(1-\alpha, 1/2 + k, 1/2 + n - k)$.%
\footnote{When $k = 0$ the lower limit is set to 0 and when $k = N$ the upper limit is set to 1 to 
avoid the coverage tending to 0 as $p$ tends to 0 or 1.}

One-sided Jeffreys intervals for $\overline{\FPR}$ and $\overline{\FNR}$ already yield narrower confidence intervals for $\empepsilon$ than previous approaches using two-sided Clopper-Pearson intervals.
For instance, an attack with 100\% accuracy over \num{2000} trials with $\delta=\num{1e-5}$ results in a $90\%$ confidence $\lb{\empepsilon}$ of \num{5.6} using two-sided CP intervals, \num{5.81} using one-sided CP intervals, and \num{6.25} using one-sided Jeffreys intervals.%
\footnote{\citet{Carlini:2021} report the first figure of \num{5.6} for \num{1000} trials, but it clearly is only achievable with \num{2000} trials.}

\subsection{Estimates from the Posterior Joint Distribution}
\label{sec:joint}

We show how to greatly improve the quality of estimates using the joint posterior of \FNRFPR to derive a credible interval for \empepsilon.
Given the probability density function $f_{\FNRFPR}$ of the joint posterior of \FNRFPR, we obtain the cumulative distribution of $\empepsilon$.

\begin{definition}[Cumulative Distribution Function of \empepsilon]
Let $\delta \in [0,1]$ and $f_{\FNRFPR}$ be the density function of the posterior joint distribution of \FNRFPR given observed counts of $\FN,\TP,\FP,\TN$ from Experiment~\ref{exp:ind-mia}.
The value of the cumulative distribution function of \empepsilon at $\epsilon$ is the integral of $f_{\FNRFPR}$ over the privacy region $\mathcal{R}(\epsilon, \delta)$:
\begin{equation}
\label{eq:cdf}
\begin{aligned}
F_{\empepsilon}(\epsilon) 
    = \Pr{(\FNR, \FPR) \in \mathcal{R}(\epsilon, \delta)}
    = \iint_{\mathcal{R}(\epsilon, \delta)} f_{\FNRFPR}(x,y)\ \dd x\ \dd y \; .
\end{aligned}
\end{equation}
\end{definition}

Equipped with $F_{\empepsilon}$ we can compute the $100(1 - \alpha)\%$ equal-tailed \emph{credible interval} $[\lb{\empepsilon},\ub{\empepsilon}]$
\begin{align}
\lb{\empepsilon} = \arg\max_\epsilon F_{\empepsilon}(\epsilon) \leq \alpha/2 
\;, \qquad
\ub{\empepsilon} = \arg\min_\epsilon F_{\empepsilon}(\epsilon) \geq 1 - \alpha/2
\label{eqn:binomial_eps}
\end{align}

The Bayesian model we presented above gives us the densities of the posteriors $\FNR|\FN$ and $\FPR|\FP$.
Since the populations of positive and negative instances are independent, it is natural to model these posteriors as independent, yielding a joint distribution we can plug into Equation~\ref{eq:cdf}:
\begin{equation*}
f_{\FNRFPR}(x,y) \eqdef f_{\FNR|\FN}(x)\ f_{\FPR|\FP}(y)
\end{equation*}
The resulting integral in Eq.~\ref{eq:cdf} cannot be expressed in analytical form so we approximate it numerically.

\begin{figure}[ht]
    \centering
    \includegraphics[width=.52\columnwidth]{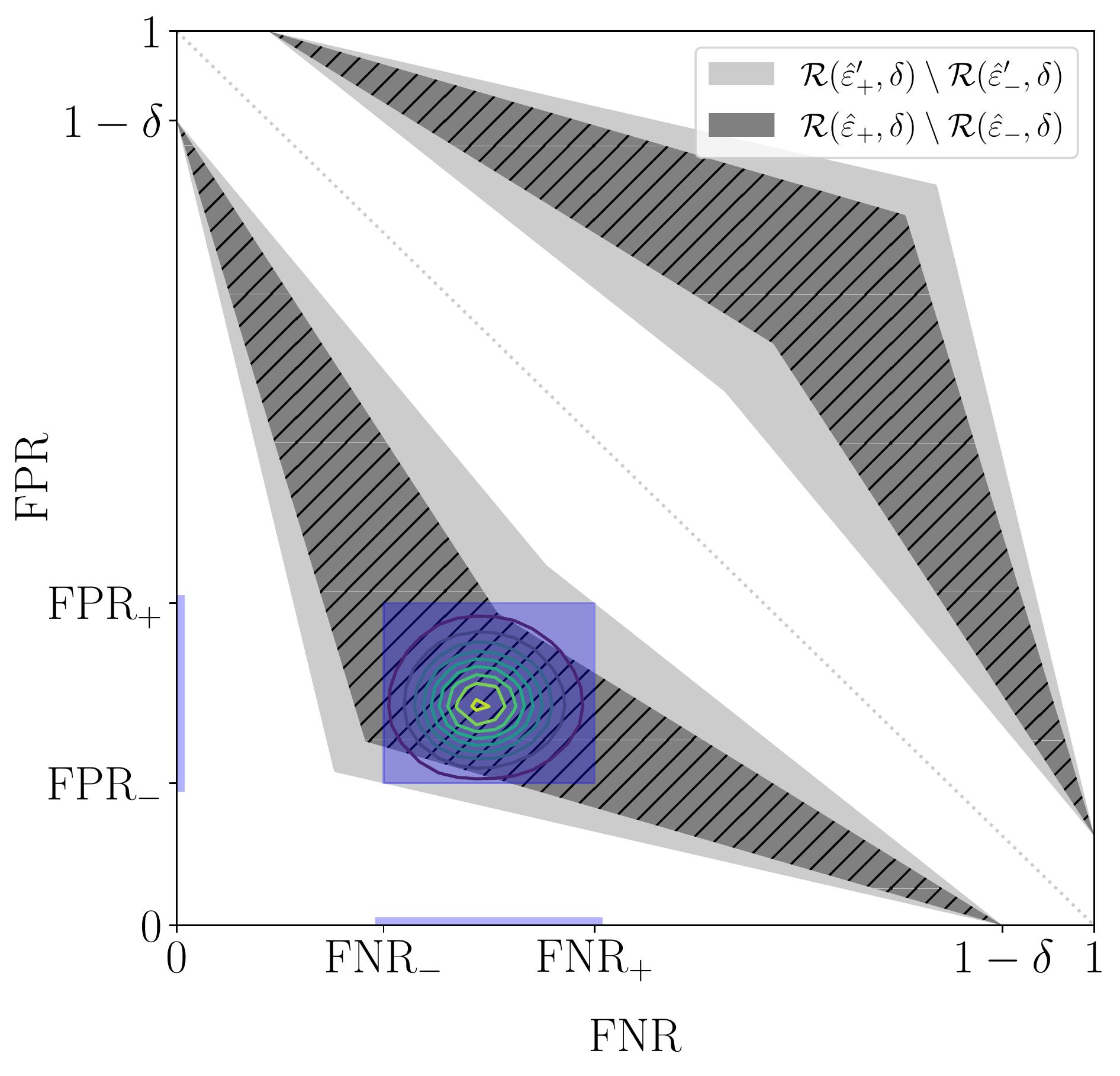}
    \caption{Graphical interpretation of intervals for \empepsilon obtained using a joint binomial model ($[\empepsilon_{-}, \empepsilon_{+}]$) and Jeffreys confidence intervals ($[\empepsilon'_{-}, \empepsilon'_{+}]$). The contour plot of the density $f_{\FNRFPR}$ and the rectangle determined by Jeffreys intervals match closely.}
    \label{fig:integral_region}
\end{figure}

Figure~\ref{fig:integral_region} provides an intuitive graphical explanation of why estimates for $\empepsilon$ derived from confidence intervals are looser than using a Bayesian approach at the same confidence level.
Taken together, confidence intervals for the false positive and false negative rate of a membership inference attack determine a rectangle in the (FNR,FPR) space.
This rectangle covers $1 - \alpha$ of the density of the joint distribution of (FNR,FPR) but fits in between two privacy regions whose difference covers strictly more density.  
A $100(1 - \alpha)\%$ confidence interval for \empepsilon derived using this method will have larger than nominal coverage because the additional density in 
$\mathcal{R}(\empepsilon'_{+}, \delta) \setminus \mathcal{R}(\empepsilon'_{-}, \delta)$ outside the rectangle is unaccounted for.
In contrast, by integrating $f_{\FNRFPR}$, we can derive a credible interval for \empepsilon with exactly the nominal coverage, barring numerical error.

For instance, suppose we run \num{200} times Experiment~\ref{exp:ind-mia}, collecting samples $\{b_i,\guess{b}_i\}$ and after tallying we get $\FN=35, \TP=65, \FP=25, \TN=75$.
To derive a $90\%$ confidence interval for \empepsilon, we compute the minimum and maximum of Eq.~\eqref{eqn:lowerbound} over the two-sided Jeffreys intervals for \FNR and \FPR obtained from the tally, which yields $[0.295,1.489]$.
To derive instead a $90\%$ credible interval, we construct the cumulative distribution function of \empepsilon by integrating $f_{\FNRFPR}$ and solve Eqs.~\eqref{eqn:binomial_eps}, which yields a narrower interval $[0.522,1.268]$.
In terms of Fig.~\ref{fig:integral_region}, the rectangle covers $96.3\%$ of the density of $f_{\FNRFPR}$, but it is enclosed in an area between two privacy regions that covers $99.8\%$. In comparison, the smaller hatched area corresponding to the Bayesian credible interval has $95\%$ coverage by definition.

%% file: parameter_eval.tex
\section{Evaluation of the Bayesian Approach}

We evaluate the performance of our Bayesian approach in numeric simulations and in experiments on text and vision classifiers. 
For both settings, we compare equal-tailed credible intervals for \empepsilon obtained using our new Bayesian approach with confidence intervals for \empepsilon derived from two-sided Clopper-Pearson and Jeffreys intervals.

\subsection{Numeric Simulation}

\paragraph{Methodology}
We assume a hypothetical attack with a fixed balanced accuracy of 60\%, from which we derive \FPR and \FNR for a given number of samples.
With this we evaluate the reduction in uncertainty by comparing confidence interval sizes for $\empepsilon$ (assuming a fixed $\delta$) based on a fixed number of samples, using Clopper-Pearson intervals, Jeffreys intervals, and our Bayesian approach.
We also evaluate the improvement in computational cost by fixing the confidence interval size and comparing the number of samples required to achieve them using the different methods.

\begin{figure*}
  \centering
  \resizebox{0.5\textwidth}{!}{\input{tikz/interval_size_comparison_060}}
  \caption{
      Credible interval size as a function of the number of samples.
      We compare the estimation techniques with $\alpha=0.1$ for an attack with 60\% balanced accuracy over varying number of challenges points. For a fixed number of samples, the distance between the matching upper and lower bounds illustrates the reduction in confidence interval size. For a fixed interval size (as illustrated by the shaded area), the difference in challenge points at which the interval bounds intersect with the shaded area illustrates the reduction in samples.
      The shaded are in the figure corresponds to a scenario where we want to estimate $\empepsilon$ within $\pm 0.15$ with a confidence of $90\%$. Our Bayesian approach reduces the number of challenge points from approximately $1,500$ to only $500$.
  }
  \label{fig:alternative_interval_size_comparison}
\end{figure*}
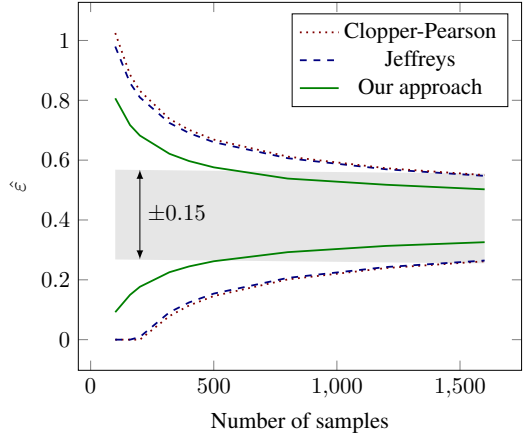

\paragraph{Results}
Figure~\ref{fig:alternative_interval_size_comparison} shows the results of this comparison.
Here we are interested in an estimate for $\empepsilon$ within $\pm 0.15$ with a significance level of $\alpha=10\%$.
The Clopper-Pearson approach requires approximately 1,500 samples.
Jeffreys intervals marginally reduce the number of samples.
Using our Bayesian approach, we can significantly reduce the number of samples to just over $500$ thereby reducing the computational cost by \nicefrac{2}{3}.

\subsection{Experiments on Text and Vision Classifiers}

We evaluate the performance of the Bayesian approach on vision and text classifiers.

\paragraph{Datasets and Tasks}
\begin{itemize} 
  \item CIFAR10~\cite{Krizhevsky:2009}, consisting of 60,000 labeled (50,000 training, 10,000 test) images containing one of ten object classes, with 6,000 images per class. We use a 4-layer CNN with 974K parameters and Tanh activations with average pooling and max pooling units, which we train for 50 epochs. Our models reach 60\% accuracy at 20 epochs and over 62\% at 50 epochs with $\epsilon=10$, $\delta=10^{-5}$. 

  \item SST 2~\cite{Socher:2013}, a binary sentiment text classification dataset consisting of 67,349 training samples and 1,821 test samples. We fine-tune a RoBERTa base model for 3 epochs to an accuracy of 92\%~\cite{Liu:2021} with $\epsilon=4$, $\delta=10^{-5}$.
\end{itemize}

\paragraph{Methodology} 
We use the false positive and false negative rates of the attacks to compute the equal-tailed confidence intervals for \empepsilon using the Clopper-Pearson and Jeffreys confidence intervals, as well as with our Bayesian approach.

\paragraph{Results} 
Table~\ref{tab:eval_bayesian} summarizes the results of this comparison on text and vision tasks using $N=1000$ samples.
Detailed results are provided in Section~\ref{sec:tables} in the Appendix.
We compute the width of confidence intervals using each method, and the reduction in interval width relative to the Clopper-Pearson method.

We observe reductions in width of between {\bf 34\%} and {\bf 40\%} for the same number of samples, demonstrating the advantage of our Bayesian approach. 
Importantly, our approach is successful in computing meaningful confidence intervals when other methods result in trivial $(0, \infty$) intervals.

\begin{table*}[ht]
\centering
\footnotesize
\setlength{\tabcolsep}{0.4em}
\renewcommand{\arraystretch}{1.3}
\begin{tabular}{M{3mm}M{3mm}M{15mm}M{16mm}M{8mm}M{16mm}M{8mm}M{9mm}M{16mm}M{8mm}M{9mm}}
\toprule
& & & \multicolumn{2}{c}{\textbf{Clopper-Pearson}} & \multicolumn{3}{c}{\textbf{Jeffreys}} & \multicolumn{3}{c}{\textbf{Bayesian Approach}}\\
\cmidrule(l){4-5} \cmidrule(l){6-8} \cmidrule(l){9-11}
& & & Interval & Width & Interval & Width & vs.\ CP & Interval & Width & vs.\ CP\\
\midrule
\multirow{2}{*}{\rotatebox[origin=c]{90}{\textbf{SST2}}}
& & No DP & (0.60, 3.3) & 2.7 & (0.69, 3.1) & 2.4 & -11\% & (1.08, 2.7) & 1.6 & \textbf{-40\%}\\
& & $\varepsilon=4$ & (0, $\infty$) & $\infty$  & (0, $\infty$) & $\infty$ & -- & (0.22, 7.0) & 6.8 & \textbf{--}\\
\cmidrule{1-11}
\multirow{4}{*}{\rotatebox[origin=c]{90}{\textbf{CIFAR10}}} 
& \multirow{2}{*}{\rotatebox[origin=c]{90}{(avg.)}}
  & No DP & (2.0, 5.8) & 3.8 & (2.1, 5.3) & 3.2 & -16\% & (2.5, 4.8) & 2.3 & \textbf{-40\%}\\
& & $\varepsilon=10$ & (0, 0.26) & 0.26 & (0, 0.25) & 0.25 & -4\% & (0.005, 0.17) & 0.16 & \textbf{-38\%}\\
\cmidrule{2-11}
& \multirow{2}{*}{\rotatebox[origin=c]{90}{(worst)}}
  & No DP & (2.9, 5.8) & 2.9  & (3.0, 5.5) & 2.5  & -13\% & (3.3, 5.2) & 1.9  & \textbf{-34\%}\\
& & $\varepsilon=10$ & (0, $\infty$) & $\infty$ & (0, $\infty$)  & $\infty$ & -- & (0.15, 6.4)  & 6.3 & \textbf{--}\\
\bottomrule
\end{tabular}
\caption{Comparison of intervals obtained from different estimation methods for text and vision models trained with and without DP ($\delta=10^{-5}$).
For CIFAR10, we compute estimates from attacks on average case and worst-case training data examples.
For each method, the bounds and widths are for equal-tailed intervals at $\alpha=0.1$.
As shown in the rightmost column, our approach provides between {\bf 34\%} to {\bf 40\%} reduction in width and can compute meaningful intervals when other approaches result in trivial $(0, \infty$) intervals.}
\label{tab:eval_bayesian}
\end{table*}

%% file: tikz/interval_size_comparison_060.tex
\begin{tikzpicture}
  \begin{axis}[
    xlabel=Number of samples,
    ylabel=$\hat\varepsilon$,
    legend pos=north east,
  ]
    \fill[gray!20]
      (100, 0.268) -- (1600, 0.255) -- (1600, 0.555) -- (100, 0.568) -- cycle ;
    \addplot[thick,black!50!red,dotted] table[x=num_samples, y=eps_beta_lo, col sep=tab] {data/epsilon-vs-num-samples-acc-060-a-10-two-sided.tsv};
    \addlegendentry{Clopper-Pearson}
    \addplot[thick,black!50!blue,dashed] table[x=num_samples, y=eps_jeffreys_lo, col sep=tab] {data/epsilon-vs-num-samples-acc-060-a-10-two-sided.tsv};
    \addlegendentry{Jeffreys}
    \addplot[thick,black!50!green] table[x=num_samples, y=eps_joint_lo, col sep=tab] {data/epsilon-vs-num-samples-acc-060-a-10-two-sided.tsv};
    \addlegendentry{Our approach}
    \addplot[thick,black!50!blue,dashed] table[x=num_samples, y=eps_jeffreys_hi, col sep=tab] {data/epsilon-vs-num-samples-acc-060-a-10-two-sided.tsv};
    \addplot[thick,black!50!red,dotted] table[x=num_samples, y=eps_beta_hi, col sep=tab] {data/epsilon-vs-num-samples-acc-060-a-10-two-sided.tsv};
    \addplot[thick,black!50!green] table[x=num_samples, y=eps_joint_hi, col sep=tab] {data/epsilon-vs-num-samples-acc-060-a-10-two-sided.tsv};
    \draw[<->] (200, 0.268) -- (200, 0.568);
    \node at (350, 0.42) {$\pm 0.15$};
  \end{axis}
\end{tikzpicture}

%% file: heuristics_eval.tex
\section{Improving Efficiency with Heuristics}
\label{sec:heuristic}

Obtaining a single sample for estimating \empepsilon requires running a MIA experiment on a model trained from scratch (see Section~\ref{sec:background}). 
This can quickly become prohibitively expensive as the sample size grows.
\citet{Malek:2021} proposed an heuristic to evaluate label-only DP that draws multiple samples from a single model.
In this section we adapt this heuristic to full DP to approximate our Bayesian approach, and perform a first analysis of its applicability. 

Our Bayesian approach enables computing the cumulative distribution function (CDF) as well as the probability density function (PDF) of \empepsilon. 
Plotting the CDFs allows us to perform a direct visual comparison between the baseline and the heuristic.

\subsection{Heuristic for Computationally-Efficient Estimation}

\begin{wrapfigure}[13]{r}{0.5\textwidth}
  \vspace{-10pt}
  \begin{minipage}{0.5\textwidth}
  \begin{experiment}[H]
    \DontPrintSemicolon
    \caption[F]{$\textrm{MIA}^m$}
         $S \gets \A_1(\mathcal{T}, \mathcal{D}, n)$ 
         \tcp*{$|S| = n-m$}
         \For{$i \gets 1$ \KwTo $m$}{
           $z_0^i, z_1^i \sim \A_2(\mathcal{T}, \mathcal{D}, n)$\;
           $b_i \sim \{0,1\}$\;
           $S \gets S \cup \{ z_{b_i}^i \}$
         }
         $\theta \gets \mathcal{T}(S)$\;
         \For{$i \gets 1$ \KwTo $m$}{
           $\guess{b}_i \gets \A_3(\mathcal{T}, \mathcal{D}, n, S \setminus \{z_{b_i}^i\}, z_0^i, z_1^i, \theta)$ 
         }
         \label{alg:heuristic}
  \end{experiment}
  \end{minipage}
\end{wrapfigure}

We formalize the heuristic as Experiment~\ref{alg:heuristic}. It resembles the MI game in Experiment~\ref{exp:ind-mia}, except that 1) the adversary creates $m$ challenge pairs, with one point from each pair chosen at random and added to the training set; and 2) the adversary {\em sequentially} receives $m$ challenge pairs and is tasked with determining which challenge point was used during training.

To obtain $N$ samples, we run $N/m$ times Experiment~\ref{alg:heuristic}, \ie, we train only $N/m$ models. We then use the approach described in Section~\ref{sec:approach} to derive estimates of \empepsilon.
Experiment~\ref{alg:heuristic} is parametric in the choices of attack and challenge points, which we instantiate next.

The crux that makes the heuristic reasonable is that in the Experiment~\ref{alg:heuristic} the attacker still gets {\bf a single} pair of samples as in Experiment~\ref{exp:ind-mia} and has to base their decision on them rather than on the $2m$ samples chosen over all iterations. The expectation is that such attacker would not gain much from the datasets differing in other, unrelated, samples from other iterations.

\paragraph{Choice of challenge points}
We adapt the heuristic from \citet{Malek:2021} by allowing the adversary to pick arbitrary challenge points (both the input and label) during the experiment, and thereby compute full DP estimates. 

Prior work crafted challenge points to maximize the signal for membership inference attacks~\cite{Jagielski:2020,Carlini:2021b}.%
\footnote{\citet{Carlini:2021b} seemingly sample challenge points at random in \S IV.A. After being unable to replicate their results we clarified by personal communication that the results they report are for worst-case samples.} In contrast, we consider an adversary that chooses natural challenge points from the population. We consider two regimes: 

\begin{inparaenum}
  \item In the {\em worst-case} regime, we select challenge points with the largest train-test loss gap, which we identify by training several models on random in/out splits of the training and validation sets. 
  
  \item In the {\em average-case} regime, we select challenge points uniformly at random from the dataset without replacement.
\end{inparaenum}

\paragraph{Choice of attack}
Our experiments use loss threshold attacks~\cite{Yeom:2018} to determine membership. Specifically, we use \emph{model-dependent} thresholds~\cite{Ye:2021}, which we choose as an $\alpha$-percentile of the empirical distribution of the losses.
The $\alpha$ value is fixed while evaluating the attack across models trained on the same dataset with different $m$ values.
We ran a linear search for $\alpha$ in the $(0,100)$ interval in a preprocessing step to find the largest lower bound estimate $\empepsilon_{-}$, based on challenge points that are chosen following the same regime (average or worst-case).
Using a global $\alpha$ to pick a different loss threshold per model results in a stronger attack than using the same threshold across all models. 
The parameter $\alpha$ can be chosen to yield the best attack for each $m$, or fixed to yield a low FPR attack.

\subsection{Evaluating the Heuristic}

Estimating \empepsilon requires many samples from independent runs of Experiment~\ref{exp:ind-mia}. However, in Experiment~\ref{alg:heuristic}, we draw $m$ samples from each of the $N/m$ models. These samples are not independent, hence the computed estimates need not be faithful when $1 < m \le N$.
We experimentally evaluate the heuristic in different scenarios to find an appropriate trade-off between $N$, the number of samples, and $N/m$, the number of models trained, and determine the limits of applicability of the heuristic.

\begin{figure}
  \centering
  \centering
  \subfloat[{{\small With DP}}]{
    \label{sst2-dp}
    \resizebox{0.45\linewidth}{!}{
      \input{tikz/heuristic_sst2-dp}
    }
  }
  \subfloat[{{\small Without DP}}]{
    \label{sst2-nodp}
    \resizebox{0.45\linewidth}{!}{
      \input{tikz/heuristic_sst2}
    }
  }
  \caption{Evaluation of heuristic on SST2 models trained with DP $\epsilon=4$ and without DP \ie, $\epsilon=\infty$. The curves correspond to CDFs of $\epsilon$ estimates for different numbers $m$ of membership queries per model.}
  \label{fig:sst2}
\end{figure}
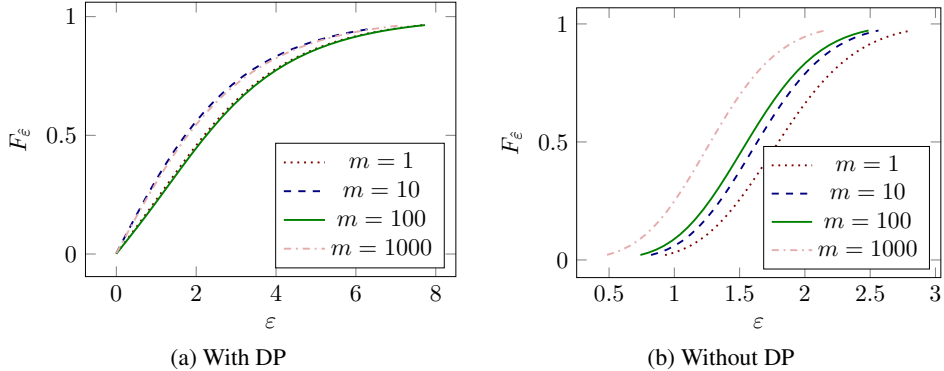

We compare the CDF of $\empepsilon$ for the baseline $m=1$ against those obtained for $m=10,100,1000$, using a fixed number of samples $N=1000$ for models trained on CIFAR10 and SST2 datasets (shown in Figure~\ref{fig:cifar10-dp} and \ref{fig:sst2}). 
We trained \num{1111} models for each setting ($1000+100+10+1$) totaling \num{6666} models using our local GPU cluster, resulting in roughly 10K GPU compute hours.

\textbf{Results.}
We evaluate the heuristics with average case and worst case challenge points for models trained with and without differential privacy.

\begin{asparaitem}
\item The results for CIFAR-10 trained with DP are given in Figure~\ref{fig:cifar10-dp}.
For the average case regime, the density functions of the estimated $\empepsilon$ lower bound values for $m=1,10,100,1000$ all coincide, with errors of $0$ for $m=10$ and $0.001$ for $m=100$ and $m=1000$, which validates the heuristic.
For the worst case regime, the CDF curves for $m=10$ and $m=100$ coincide with the baseline with an absolute error of $0.05$ in both cases.
However, we observe that using only a single model ($m=\num{1000}$) introduces a large error, indicating a decrease in performance on worst case challenge points.

\item The results for SST2 trained with DP are given in Figure~\ref{sst2-dp}.
Our results show loosely matching CDF curves for all values of $m$, with a maximum error of $0.07$. 
Similar to CIFAR-10, this allows us to reduce the computation overhead required for estimating DP lower bounds by 3 orders of magnitude.

\item Our proposed approach does not require models to be trained with differential privacy. Therefore, we can estimate $\empepsilon$ for vanilla models trained without DP, which we do for SST2 without DP in Figure~\ref{sst2-nodp}. Observe that the $m=10$ curve is the closest to the baseline of $m=1$ with an error of $0.12$. Since this is larger than for models trained with DP, we conclude that the heuristic does not perform as well for computing estimates of non-DP models.
\end{asparaitem}

\begin{figure}
  \centering
  \subfloat[{{\small Average Case }}]{
    \label{avg-case-cifar10-dp}
    \resizebox{0.45\linewidth}{!}{
      \input{tikz/heuristic_cifar_dp_avg}
    }
  }
  \subfloat[{{\small Worst Case }}]{
    \label{worst-case-cifar10-dp}
    \resizebox{0.45\linewidth}{!}{
      \input{tikz/heuristic_cifar_dp_worst}
    }
  }
  \caption{Evaluation of heuristic on CIFAR10 models trained with $\epsilon=10$. The curves correspond to CDFs of $\empepsilon$ for different values of $m$, the number of samples drawn per model.}
  \label{fig:cifar10-dp}
\end{figure}
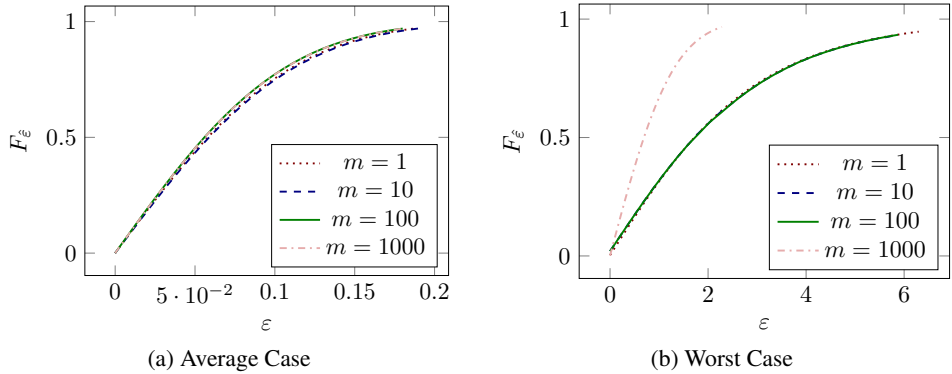

\paragraph{Limitations}
We have not evaluated the heuristic on models trained with significantly lower or higher DP $\epsilon$ values, and therefore cannot make claims about how it would perform in those regions.
We also cannot comment on how it might perform when using weaker membership inference attacks.

\paragraph{Summary}
Our results show that the heuristic generally yields faithful estimates for models trained with DP. We get two to three orders of improvement in computation cost based on the selection of challenge points. On models trained {\em without} DP, the heuristic leads to significant under-estimation of the empirical DP bounds compared to the $m=1$ baseline.

%% file: tikz/heuristic_cifar_dp_avg.tex
\begin{tikzpicture}
  \begin{axis}[
    xlabel=$\varepsilon$,
    ylabel=$F_{\hat\varepsilon}$,
    legend pos=south east,
    height=0.4\textwidth,
    width=0.5\textwidth,
  ]
    \addplot[thick,black!50!red,dotted] table[x=epsilon, y=pdf, col sep=tab] {data/CIFAR+DP-average-m_1.tsv};
    \addlegendentry{$m=1$}
    \addplot[thick,black!50!blue,dashed] table[x=epsilon, y=pdf, col sep=tab] {data/CIFAR+DP-average-m_10.tsv};
    \addlegendentry{$m=10$}
    \addplot[thick,black!50!green] table[x=epsilon, y=pdf, col sep=tab] {data/CIFAR+DP-average-m_100.tsv};
    \addlegendentry{$m=100$}
    \addplot[thick,black!10!pink,dash dot] table[x=epsilon, y=pdf, col sep=tab] {data/CIFAR+DP-average-m_1000.tsv};
    \addlegendentry{$m=1000$}
  \end{axis}
\end{tikzpicture}

%% file: tikz/heuristic_cifar_dp_worst.tex
\begin{tikzpicture}
  \begin{axis}[
    xlabel=$\varepsilon$,
    ylabel=$F_{\hat\varepsilon}$,
    legend pos=south east,
    height=0.4\textwidth,
    width=0.5\textwidth,
  ]
    \addplot[thick,black!50!red,dotted] table[x=epsilon, y=pdf, col sep=tab] {data/CIFAR+DP-worst-m_1.tsv};
    \addlegendentry{$m=1$}
    \addplot[thick,black!50!blue,dashed] table[x=epsilon, y=pdf, col sep=tab] {data/CIFAR+DP-worst-m_10.tsv};
    \addlegendentry{$m=10$}
    \addplot[thick,black!50!green] table[x=epsilon, y=pdf, col sep=tab] {data/CIFAR+DP-worst-m_100.tsv};
    \addlegendentry{$m=100$}
    \addplot[thick,black!10!pink,dash dot] table[x=epsilon, y=pdf, col sep=tab] {data/CIFAR+DP-worst-m_1000.tsv};
    \addlegendentry{$m=1000$}
  \end{axis}
\end{tikzpicture}

%% file: tikz/heuristic_sst2-dp.tex
\begin{tikzpicture}
  \begin{axis}[
    xlabel=$\varepsilon$,
    ylabel=$F_{\hat\varepsilon}$,
    legend pos=south east,
    height=0.4\textwidth,
    width=0.5\textwidth,
  ]
    \addplot[thick,black!50!red,dotted] table[x=epsilon, y=pdf, col sep=tab] {data/SST2+DP-m_1.tsv};
    \addlegendentry{$m=1$}
    \addplot[thick,black!50!blue,dashed] table[x=epsilon, y=pdf, col sep=tab] {data/SST2+DP-m_10.tsv};
    \addlegendentry{$m=10$}
    \addplot[thick,black!50!green] table[x=epsilon, y=pdf, col sep=tab] {data/SST2+DP-m_100.tsv};
    \addlegendentry{$m=100$}
    \addplot[thick,black!10!pink,dash dot] table[x=epsilon, y=pdf, col sep=tab] {data/SST2+DP-m_1000.tsv};
    \addlegendentry{$m=1000$}
  \end{axis}
\end{tikzpicture}

%% file: tikz/heuristic_sst2.tex
\begin{tikzpicture}
  \begin{axis}[
    xlabel=$\varepsilon$,
    ylabel=$F_{\hat\varepsilon}$,
    legend pos=south east,
    height=0.4\textwidth,
    width=0.5\textwidth,
  ]
    \addplot[thick,black!50!red,dotted] table[x=epsilon, y=pdf, col sep=tab] {data/SST2-DP-m_1.tsv};
    \addlegendentry{$m=1$}
    \addplot[thick,black!50!blue,dashed] table[x=epsilon, y=pdf, col sep=tab] {data/SST2-DP-m_10.tsv};
    \addlegendentry{$m=10$}
    \addplot[thick,black!50!green] table[x=epsilon, y=pdf, col sep=tab] {data/SST2-DP-m_100.tsv};
    \addlegendentry{$m=100$}
    \addplot[thick,black!10!pink,dash dot] table[x=epsilon, y=pdf, col sep=tab] {data/SST2-DP-m_1000.tsv};
    \addlegendentry{$m=1000$}
  \end{axis}
\end{tikzpicture}

%% file: related.tex
\section{Related Work}
\label{sec:related}

\textbf{Empirical Privacy Estimates.}
%
\citet{Hyland:2019} estimate DP bounds based on an empirical estimate of the sensitivity of SGD. 
\citet{Jagielski:2020} derive estimates from black-box membership inference attacks, using clipping-aware poisoning attacks against DP-SGD.
\citet{Carlini:2021b} use similar techniques but consider a hierarchy of adversaries, ranging from black-box membership inference to distinguishers that craft worst-case datasets.
Both works derive estimates from Clopper-Pearson confidence intervals of the false positive and false negative rates of attacks. 
Our Bayesian approach is generally applicable in the same settings and consistently yields tighter estimates for the same number of samples.

\textbf{DP violations.}
Several approaches~\cite{Ding:2018,Bichsel:2021} find violations of DP claims by constructing counterexamples (\ie, adjacent inputs together with a distinguishing test). These approaches aim to falsify a conjectured guarantee, whereas we aim to estimate an unknown guarantee for a given threat model. More fundamentally, these approaches are applicable to DP mechanisms beyond ML training but require the search space to be sufficiently constrained for the counterexample search to succeed. In contrast, we compute estimates with respect to a given class of parametrized distinguishers which allows us to run a much more efficient search over relatively small parameter space.

\textbf{Membership Inference attacks.}
Our approach is parametric on the choice of membership inference attack. Early membership inference attacks relied on training shadow models~\cite{Shokri:2017}. Threshold-based attacks were introduced by~\cite{Yeom:2018}. \citet{Ye:2021} compare different strategies to choose loss thresholds. In our evaluation, we choose model-dependent thresholds as they offer an attractive trade-off between accuracy and computational cost. \citet{Carlini:2022} challenge the use of attack accuracy as a meaningful way to evaluate empirical privacy and instead propose to measure false positive rates at low false negative rates. Our evaluation shows that our Bayesian approach performs particularly well in this regime. It also obtains meaningful estimates where prior approaches would result in intervals including 0 and the known theoretical bound (see \eg, Table~\ref{tab:sst_dp}).
\citet{Yaghini:2022} show that different cohorts of samples can exhibit disparate vulnerability to membership inference and prove that differential privacy bounds the magnitude of the disparity. It would be interesting to study how this disparity correlates with empirical estimates of differential privacy.

\textbf{Provable DP Bounds.}
Since the introduction of DP-SGD and the Moments Accountant technique~\cite{Abadi:2016}, there have been steady improvements in privacy accounting techniques, resulting in tighter and tighter privacy budget accounting for DP-SGD. However, this trend cannot continue as state-of-the-art accountants are tight~\cite{Gopi:2021}. Further improvements require different algorithms such as PATE~\cite{Papernot:2017}, or the introduction of additional assumptions such as weaker adversary models or convexity~\cite{Chourasia:2021}.

%% file: conclusion.tex
\section{Conclusion}

We propose a novel Bayesian approach that yields high-confidence estimates of bounds on the differential privacy parameter $\epsilon$ of training pipelines.
We show experimentally that our approach, combined with a heuristic from prior work on label-only DP that we adapt and validate, translates into privacy estimates that are tighter than using existing approaches and that can be obtained at a fraction of the computational cost.

%% file: intervals.tex
\section{Privacy Estimates Derived from Confidence Intervals}
\label{sec:montecarlo}

Given samples $\{b_i,\guess{b}_i\}$ from runs of Experiment~\ref{exp:ind-mia}, we compute sample estimates and intervals for \FPR and \FNR:

\begin{align*}
\overline{\FPR} &= \frac{\sum_{i=1}^{m} \left[ \guess{b}_i \neq b_i \land b_i = 0 \right]}
                        {\sum_{i=1}^{m} \left[ b_i = 0 \right]} 
                        \in [\lb{\FPR}, \ub{\FPR}] \\
\qquad
\overline{\FNR} &= \frac{\sum_{i=1}^{m} \left[ \guess{b}_i \neq b_i \land b_i = 1 \right]}
                        {\sum_{i=1}^{m} \left[ b_i = 1 \right]}
                        \in [\lb{\FNR}, \ub{\FNR}]
\end{align*}

A lower bound for \empepsilon can be computed minimizing Eq.~\eqref{eqn:lowerbound} over these confidence intervals (where the terms are well-defined).
\cite[Eq. 5]{Carlini:2021} simply take the value at $(\ub{\FNR}, \ub{\FPR})$, but special care should be taken when either \lb{\FPR} or \lb{\FNR} is 0 as the minimum can occur at \eg, $(\ub{\FNR}, 0)$.  
An upper bound $\ub{\empepsilon}$ can be computed analogously, but is less interesting since it does not bound the privacy afforded by the training pipeline \wrt more powerful adversaries. 

From the union bound, the significance of the confidence interval for $\empepsilon$ is double the significance of the confidence intervals for $\overline{\FPR}$ and $\overline{\FNR}$ used to derive it.
For instance, when using $95\%$ confidence intervals for $\overline{\FPR}$ and $\overline{\FNR}$, the derived confidence interval $[\lb{\empepsilon}, \ub{\empepsilon}]$ has $90\%$ confidence. 

For example, when TP, FP, TN, FN = (90, 0, 100, 10) and $\delta = \num{1e-5}$ using 90\% Clopper-Pearson intervals, the value of Eq.~\ref{eqn:lowerbound} at $(\ub{\FNR}, \ub{\FPR})$ is \num{3.124}, while the minimum occurs at $(\ub{\FNR}, 0)$ and is \num{1.736}.

\subsection{About Clopper-Pearson Confidence Intervals}

Sample false negative (\FN) and false positive counts (\FP) can be modeled as the number of successes of two binomial distributions with respective unknown success probabilities \FNR and \FPR. 
Given $k$ observed successes in $N$ trials, the lower and upper limits of the two-sided $100(1 - \alpha)\%$ Clopper-Pearson interval are respectively the solutions $p$ to the equations $\Pr{\Bin(N, p) \geq k} = \alpha/2$ and $\Pr{\Bin(N, p) \leq k} = \alpha/2$. 
The interval can be succinctly written in terms of quantiles of Beta distributions as $[\BetaPPF(\alpha/2, k, N - k +1), \BetaPPF(1-\alpha/2, k + 1, N - k)]$, where $\BetaPPF(q, a, b)$ is the $q$ quantile of $\Beta(a, b)$. 

Clopper-Pearson intervals are guaranteed to have at least their nominal coverage. However, they typically exceed it, which results in privacy estimates that are overly conservative.
An obvious improvement over the state-of-the-art approach to lower bound $\empepsilon$ \citep{Carlini:2021} is to use one-sided Clopper-Pearson intervals since \lb{\empepsilon} only depends on their upper-limit (\lb{\FPR} and \lb{\FNR} are only 0 when \FP or \FN are exactly 0). This effectively doubles the significance of estimates.

An alternative to exact Clopper-Pearson intervals are \emph{approximate} confidence intervals, which have coverage closer to nominal, such as Jeffreys intervals.

%% file: mia.tex
\section{Relation between Differential Privacy and Membership Inference}

\citet{Yeom:2018} formalize membership inference attacks with balanced priors as a game equivalent to Experiment~\ref{exp:mia}.


\begin{experiment}[h!]
  \DontPrintSemicolon
  \caption[F]{MIA}
    \KwIn{$\mathcal{T}, \mathcal{D}, n, \A$}
    $S \sim \mathcal{D}^{n-1}$; $z_0, z_1 \sim \mathcal{D}^2$\;
    $b \sim \{0,1\}$\;
    $\theta \gets \mathcal{T}(S \cup \{z_b\})$\; 
    $\guess{b} \gets \A(\mathcal{T}, \mathcal{D}, n, \theta, z_0)$
\label{exp:mia}
\end{experiment}

\begin{definition}[Membership Inference Advantage]
\label{def:mia} 
The membership inference advantage $\Adv{MIA}(\mathcal{T}, \mathcal{D}, n, \A)$ of adversary $\A$ is the quantity
\begin{equation*}
   2\,\Pr{\mathrm{MIA}(\mathcal{T}, \mathcal{D}, n, \A) : \guess{b} = b} - 1
\end{equation*}
\end{definition}

\begin{figure}[ht]
\centering
\includegraphics[width=0.5\textwidth]{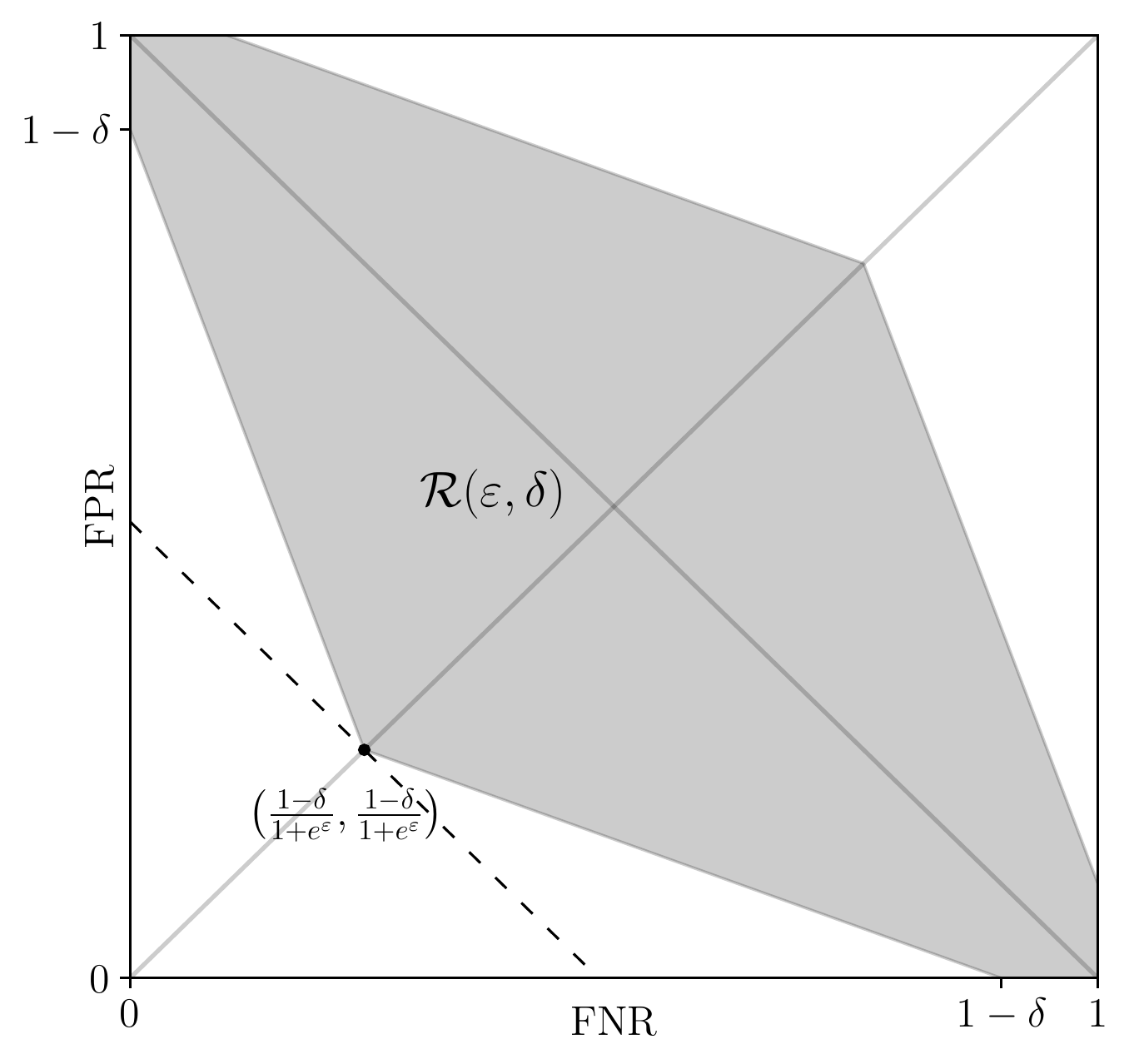}
\caption{Privacy region $\mathcal{R}(\epsilon,\delta)$.
The dashed line corresponds to the bound of Theorem~\ref{thm:mia_bound} and intersects the region at the marked vertex. Previous, loose bounds correspond to parallel lines that do not intersect $\mathcal{R}(\epsilon,\delta)$.
}
\label{fig:privacy_region_mia}
\end{figure}

Interestingly, $\Adv{MIA}(\mathcal{T}, \mathcal{D}, n, \A) = (1 - \FNR) - \FPR$, which suggests a graphical interpretation of the bound in Theorem~\ref{thm:mia_bound}. 
The line $\FNR = 1 - \FPR - (e^{\epsilon} - 1 + 2\delta)(e^{\epsilon} + 1)^{-1}$ intersects the privacy region at the vertex marked in Figure~\ref{fig:privacy_region_mia}. 
The false negative and false positive rates at this vertex need not be achievable by any membership inference adversary, but there exist cases where they are, meaning that the bound of Theorem~\ref{thm:mia_bound} is tight. 

\citet{Humphries:2021} prove the following upper bound on the membership advantage against $(\epsilon,\delta)$-differentially private training 
algorithms, which improves over previous bounds~\citep{Erlingsson:2019}.
This bound holds for adversaries more informed than in Experiment~\ref{exp:mia} who can observe the dataset $S$ and the challenge points $z_0,z_1$. In fact, the bound holds for general DP distinguishers that not only observe these values but \emph{choose} them, as in Experiment~\ref{exp:ind-mia} in Section~\ref{sec:background}.

\begin{theorem}
\label{thm:mia_bound}
Let $\mathcal{T} : X^n \to \Theta$ be $(\epsilon,\delta)$-differentially private. Then, for any adversary $\A$,
\begin{equation*}
  \Adv{\mathrm{MIA}}(\mathcal{T}, \mathcal{D}, n, \A) \leq 
  \frac{e^{\epsilon} - 1 + 2\delta}{e^{\epsilon} + 1}
\end{equation*}
\end{theorem}

%% file: density.tex
\section{Probability Density Function of \empepsilon}
\label{sec:appendix}

We describe here how to derive a probability density function for \empepsilon.
The derivative of the cumulative distribution function $F_{\empepsilon}$ is given by
\begin{equation}
\begin{aligned}
\hat f_{\varepsilon}(\varepsilon) &= \frac{\dd}{\dd \varepsilon} F_{\varepsilon} (\varepsilon)
                                  = \frac{\dd}{\dd \varepsilon} \iint_{\mathcal{R}(\varepsilon, \delta)} f_{\FNRFPR}(x,y) \dd x \dd y 
                                  = \oint_{\partial \mathcal{R}(\varepsilon, \delta)} f_{\FNRFPR} \bm{v}_{\varepsilon} \cdot \bm{n} \dd L  
                                  \label{eqn:pdf_eps_1}
\end{aligned}
\end{equation}
where we have used Reynolds transport theorem in the last equation.

The symbol $\bm{v}_{\varepsilon}$ denotes the derivative of the boundary with respect to $\varepsilon$ and $\bm{n}$ is the outward pointing normal vector of a boundary element.

In order to make this more concrete, let us parameterize the boundary of the privacy region using the following curves
\begin{align*}
    \bm{R}_{\lo}(\varepsilon, \delta, x) &:= 
    \begin{bmatrix}
        x \\
        \max \left \{ 0, 1-\delta-\eeps x, (1-\delta-x)\textnormal{e}^{-\varepsilon} \right \}
    \end{bmatrix} \;  \\
    \bm{R}_{\hi}(\varepsilon, \delta, x) &:= 
    \begin{bmatrix}
        x \\
        \min \left \{ 1, (\delta-x)\textnormal{e}^{-\varepsilon}, \delta + (1-x)\eeps \right \}
    \end{bmatrix} \; .
\end{align*}

Note that $\partial \mathcal{R}(\varepsilon, \delta) = \bm{R}_{\lo}(\varepsilon, \delta, [0,1]) \cup \bm{R}_{\hi}(\varepsilon, \delta, [0,1])$.

Applying this to compute $\bm{v}_{\varepsilon}$ and $\bm{n}$ from Eq.~\ref{eqn:pdf_eps_1} yields
\begin{align}
    \bm{v}_{\lo} &= \partial_{\varepsilon} \bm{R}_{\lo}(\varepsilon, \delta, x) \; , \\
    \bm{n}_{\lo} &= \frac{Q \partial_x \bm{R}_{\lo}(\varepsilon, \delta, x)}{\| \partial_x \bm{R}_{\lo}(\varepsilon, \delta, x) \|} \; ,
\end{align}
where $Q$ denotes a rotation matrix performing a clockwise rotation by $\pi/2$.

Similarly, we have
\begin{align}
    \bm{v}_{\hi} &= \partial_{\varepsilon} \bm{R}_{\hi}(\varepsilon, \delta, x) \; , \\
    \bm{n}_{\hi} &= \frac{Q \partial_{-x} \bm{R}_{\hi}(\varepsilon, \delta, x)}{\| \partial_x \bm{R}_{\lo}(\varepsilon, \delta, x) \|} \; .
\end{align}

We can then plug this expression into equation (\ref{eqn:pdf_eps_1}).
Splitting the closed line integral into an integral over the upper and lower path gives
\begin{equation}
    \hat f_{\varepsilon}(\varepsilon) = 
    \int_0^1 f_{\FNRFPR}(\bm{R}_{\lo}(\varepsilon, \delta, x)) \bm{v}_{\lo} \cdot \bm{n}_{\lo} \dd x + 
    \int_1^0 f_{\FNRFPR}(\bm{R}_{\hi}(\varepsilon, \delta, x)) \bm{v}_{\hi} \cdot \bm{n}_{\hi} \dd x \; .
\end{equation}

Note, however that $\hat f_{\varepsilon}$ is not a probability density function since it is not normalized.
The mass of the privacy region for $\varepsilon=0$ is missing: $\int_0^{\infty} f_{\varepsilon}(\varepsilon) \dd \varepsilon = 1-F_{\varepsilon}(0) \neq 1 $.
We can correct for that by adding a point mass at $\varepsilon=0$ which gives a final expression for the probability density of $\varepsilon$
\begin{equation}
    f_{\varepsilon}(\varepsilon) = F_{\varepsilon}(0)\delta(\varepsilon) + \hat f_{\varepsilon}(\varepsilon) \; ,
\end{equation}
where $\delta$ is the Dirac $\delta$ distribution.

%% file: tables.tex
\section{Evaluation of the Bayesian Approach -- Omitted Results}
\label{sec:tables}

We show results omitted in the body of the paper in \cref{tab:sst_nodp,tab:sst_dp,tab:cifar_ac_nodp,tab:cifar_wc_nodp,tab:cifar_wc_dp,tab:cifar_ac_dp}.
Observe that our Bayesian approach consistently outperforms previous approaches based on confidence intervals, and in some cases (\eg Table~\ref{tab:sst_dp}) succeed to compute meaningful bounds where previous approaches report trivial intervals.

\begin{table*}[ht]
\centering
\begin{tabular}{@{}r@{~~}c@{~}c@{~}c@{~}c@{~~~}c@{~~}c@{~~}r@{\quad}c@{~~}c@{~~}r@{\quad}c@{~}c@{}}
\toprule
\multicolumn{5}{c}{}             & \multicolumn{3}{c}{Bayesian}         & \multicolumn{3}{c}{Jeffreys} & \multicolumn{2}{c}{Clopper-Pearson} \\
\cmidrule(r){6-8} \cmidrule(r){9-11} \cmidrule{12-13}
m    & TP   & TN   & FP   & FN   & Interval    & Width & vs CP          & Interval    & Width & vs CP  & Interval    & Width \\
\midrule
1    & 31   & 996  & 5    & 968  & (1.08, 2.7) & 1.6   & \textbf{-40\%} & (0.69, 3.1) & 2.4   & -11\%  & (0.60, 3.3) & 2.7 \\
10   & 31   & 1002 & 6    & 961  & (0.96, 2.4) & 1.4   & \textbf{-44\%} & (0.58, 2.9) & 2.3   & -8\%   & (0.50, 3.0) & 2.5 \\
100  & 30   & 979  & 6    & 985  & (0.88, 2.3) & 1.4   & \textbf{-44\%} & (0.49, 2.8) & 2.3   & -8\%   & (0.42, 2.9) & 2.5 \\
1000 & 25   & 1004 & 7    & 964  & (0.62, 2.0) & 1.4   & \textbf{-44\%} & (0.22, 2.5) & 2.3   & -8\%   & (0.14, 2.6) & 2.5 \\ 
\bottomrule
\end{tabular}
\caption{Comparison of intervals given by estimation methods for RoBERTa trained on SST2 without DP. 
For each method, we present the bounds and widths for the equal-tailed intervals at $\alpha=0.1$.}
\label{tab:sst_nodp}
\end{table*}

\begin{table*}[ht]
\centering
\begin{tabular}{@{}r@{~~}c@{~}c@{~}c@{~}c@{~~~}c@{~~}c@{~~}r@{\quad}c@{~~}c@{~~}r@{\quad}c@{~}c@{}}
\toprule
\multicolumn{5}{c}{}             & \multicolumn{3}{c}{Bayesian}& \multicolumn{3}{c}{Jeffreys}     & \multicolumn{2}{c}{Clopper-Pearson} \\
\cmidrule(r){6-8} \cmidrule(r){9-11} \cmidrule{12-13}
m    & TP   & TN   & FP   & FN   & Interval    & Width & vs CP & Interval      & Width    & vs CP & Interval      & Width \\
\midrule
1    & 2    & 487  & 0    & 511  & (0.22, 7.0) & 6.8   & ---   & (0, $\infty$) & $\infty$ & ---   & (0, $\infty$) & $\infty$ \\
10   & 0    & 510  & 1    & 489  & (0.15, 6.4) & 6.3   & ---   & (0, $\infty$) & $\infty$ & ---   & (0, $\infty$) & $\infty$ \\
100  & 2    & 501  & 0    & 497  & (0.23, 7.0) & 6.8   & ---   & (0, $\infty$) & $\infty$ & ---   & (0, $\infty$) & $\infty$ \\
1000 & 1    & 511  & 0    & 488  & (0.15, 6.5) & 6.35  & ---   & (0, $\infty$) & $\infty$ & ---   & (0, $\infty$) & $\infty$ \\
\bottomrule
\end{tabular}
\caption{Comparison of intervals given by estimation methods for RoBERTa trained on SST2 with $(4, 10^{-5})$-DP.
For each method, we present the bounds and widths for the equal-tailed intervals at $\alpha=0.1$.
In all cases, the Bayesian approach yields a meaningful interval whereas Jeffreys and Clopper-Pearson intervals are trivial.}
\label{tab:sst_dp}
\end{table*}

\begin{table*}[ht]
\centering
\begin{tabular}{@{}r@{~~}c@{~}c@{~}c@{~}c@{~~~}c@{~~}c@{~~}r@{\quad}c@{~~}c@{~~}r@{\quad}c@{}c@{}}
\toprule
\multicolumn{5}{c}{}             & \multicolumn{3}{c}{Bayesian}        & \multicolumn{3}{c}{Jeffreys}  & \multicolumn{2}{c}{Clopper-Pearson} \\
\cmidrule(r){6-8} \cmidrule(r){9-11} \cmidrule{12-13}
m    & TP   & TN   & FP   & FN   & Interval   & Width & vs CP          & Interval    & Width & vs CP  & Interval    & Width \\
\midrule
1    & 63   & 511  & 2    & 424  & (2.5, 4.8) & 2.3   & \textbf{-40\%} & (2.1, 5.3)  & 3.2   & -16\%  & (2.0, 5.8)  & 3.8 \\
10   & 84   & 503  & 5    & 408  & (2.2, 3.6) & 1.4   & \textbf{-46\%} & (1.9, 4.0)  & 2.1   & -19\%  & (1.8, 4.2)  & 2.6 \\
100  & 39   & 513  & 3    & 445  & (1.7, 3.7) & 2.0   & \textbf{-39\%} & (1.4, 4.2)  & 2.8   & -15\%  & (1.2, 4.5)  & 3.3 \\
1000 & 21   & 534  & 2    & 443  & (1.4, 3.8) & 2.4   & \textbf{-44\%} & (0.90, 4.5) & 3.6   & -16\%  & (0.74, 5.0) & 4.3 \\
\bottomrule
\end{tabular}
\caption{Comparison of intervals given by estimation methods for CNN trained on CIFAR10 (average case) without DP. Bounds and widths are presented for equal-tailed intervals at $\alpha=0.1$.}
\label{tab:cifar_ac_nodp}
\end{table*}

\begin{table*}[ht]
\centering
\begin{tabular}{@{}r@{~~}c@{~}c@{~}c@{~}c@{~~~}c@{~~}c@{~~}r@{\quad}c@{~~}c@{~~}r@{\quad}c@{}c@{}}
\toprule
\multicolumn{5}{c}{}             & \multicolumn{3}{c}{Bayesian}           & \multicolumn{3}{c}{Jeffreys} & \multicolumn{2}{c}{Clopper-Pearson} \\
\cmidrule(r){6-8} \cmidrule(r){9-11} \cmidrule{12-13}
m    & TP   & TN   & FP   & FN   & Interval      & Width & vs CP          & Interval   & Width & vs CP   & Interval   & Width \\
\midrule
1    & 179  & 510  & 3    & 308  & (3.3, 5.2)    & 1.9   & \textbf{-34\%} & (3.0, 5.5) & 2.5   & -13\%   & (2.9, 5.8) & 2.9  \\
10   & 103  & 416  & 92   & 389  & (0.015, 0.36) & 0.35  & \textbf{-31\%} & (0, 0.50)  & 0.5   & -2\%    & (0, 0.51)  & 0.51 \\
100  & 81   & 451  & 65   & 403  & (0.052, 0.54) & 0.49  & \textbf{-32\%} & (0, 0.71)  & 0.71  & -2.7\%  & (0, 0.73)  & 0.73 \\
1000 & 82   & 455  & 81   & 382  & (0.016, 0.39) & 0.37  & \textbf{-35\%} & (0, 0.55)  & 0.55  & -3.5\%  & (0, 0.57)  & 0.57 \\
\bottomrule
\end{tabular}
\caption{Comparison of intervals given by estimation methods for CNN trained on CIFAR10 (worst case) without DP. 
Bounds and widths are presented for equal-tailed intervals at $\alpha=0.1$.}
\label{tab:cifar_wc_nodp}
\end{table*}

\begin{table*}[ht]
\centering
\begin{tabular}{@{}r@{~~}c@{~}c@{~}c@{~}c@{~~}c@{~~}c@{~~}r@{\quad}c@{~~}c@{~~}r@{\quad}c@{~}c@{}}
\toprule
\multicolumn{5}{c}{}             & \multicolumn{3}{c}{Bayesian}          & \multicolumn{3}{c}{Jeffreys}      & \multicolumn{2}{c}{Clopper-Pearson} \\
\cmidrule(r){6-8} \cmidrule(r){9-11} \cmidrule{12-13}
m    & TP   & TN   & FP   & FN   & Interval     & Width & vs CP          & Interval      & Width     & vs CP & Interval      & Width \\
\midrule
1    & 487  & 1    & 512  & 0    & (0.15, 6.4)  & 6.3   & ---            & (0, $\infty$) & $\infty$  & ---   & (0, $\infty$) & $\infty$ \\
10   & 492  & 0    & 508  & 0    & (0.098, 5.9) & 5.8   & ---            & (0, $\infty$) & $\infty$  & ---   & (0, $\infty$) & $\infty$ \\
100  & 484  & 0    & 516  & 0    & (0.098, 6.0) & 5.9   & ---            & (0, $\infty$) & $\infty$  & ---   & (0, $\infty$) & $\infty$ \\
1000 & 461  & 2    & 534  & 3    & (0.062, 2.1) & 2.0   & \textbf{-46\%} & (0, 3.1)      & 3.1       & -16\% & (0, 3.7)      & 3.7 \\
\bottomrule
\end{tabular}
\caption{Comparison of intervals given by estimation methods for CNN trained on CIFAR10 (worst case) $(10, 10^{-5}$)-DP.
Bounds and widths are presented for equal-tailed intervals at $\alpha=0.1$.}
\label{tab:cifar_wc_dp}
\end{table*}

\begin{table*}[ht]
\centering
\begin{tabular}{@{}r@{~~}c@{~}c@{~}c@{~}c@{~~~}c@{~~}c@{~~}r@{~~~}c@{~~}c@{~~}r@{~~~}c@{~}c@{}}
\toprule
\multicolumn{5}{c}{}             & \multicolumn{3}{c}{Bayesian}            & \multicolumn{3}{c}{Jeffreys} & \multicolumn{2}{c}{Clopper-Pearson} \\
\cmidrule(r){6-8} \cmidrule(r){9-11} \cmidrule{12-13}
m    & TP   & TN   & FP   & FN   & Interval       & Width & vs CP          & Interval  & Width & vs CP    & Interval    & Width \\
\midrule
1    & 175  & 325  & 188  & 312  & (0.0054, 0.17) & 0.16  & \textbf{-38\%} & (0, 0.25) & 0.25  & -4\%     & (0, 0.26)   & 0.26 \\
10   & 173  & 326  & 182  & 319  & (0.0055, 0.17) & 0.16  & \textbf{-38\%} & (0, 0.26) & 0.26  & 0\%      & (0, 0.26)   & 0.26 \\
100  & 183  & 317  & 199  & 301  & (0.0052, 0.16) & 0.15  & \textbf{-40\%} & (0, 0.24) & 0.24  & -4\%     & (0, 0.25)   & 0.25 \\
1000 & 177  & 327  & 209  & 287  & (0.0052, 0.16) & 0.15  & \textbf{-40\%} & (0, 0.24) & 0.25  & -4\%     & (0, 0.25)   & 0.25 \\
\bottomrule
\end{tabular}
\caption{Comparison of intervals given by estimation methods for CNN trained on CIFAR10 (average case) $(10, 10^{-5}$)-DP. 
Bounds and widths are presented for equal-tailed intervals at $\alpha=0.1$.}
\label{tab:cifar_ac_dp}
\end{table*}

%% file: convergence.tex
\section{Illustration of Convergence of the Joint Posterior}

For this illustration we find an interval of possible values of $\varepsilon$ in which the true $\varepsilon$ lies with a given probability.
For convenience, we define the two-sided privacy region $\tilde{\mathcal{R}}$ as follows
\begin{equation}
\tilde{\mathcal{R}}(\varepsilon_{-}, \varepsilon_{+}, \delta) := 
  \mathcal{R}(\varepsilon_{+}, \delta) \setminus \mathcal{R}(\varepsilon_{-}, \delta) \; .
\end{equation}

\begin{figure}[ht]
  \centering
  \subfloat[$\Pr{\tilde{\mathcal{R}}(0.0051, 3.1, 0.01)=0.95}$]{\includegraphics[width=0.48\textwidth]{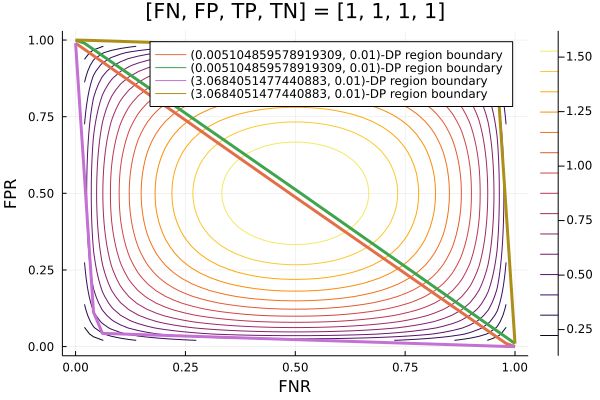}}
  \subfloat[$\Pr{\tilde{\mathcal{R}}(0.15, 3.5, 0.01)=0.95}$]{\includegraphics[width=0.48\textwidth]{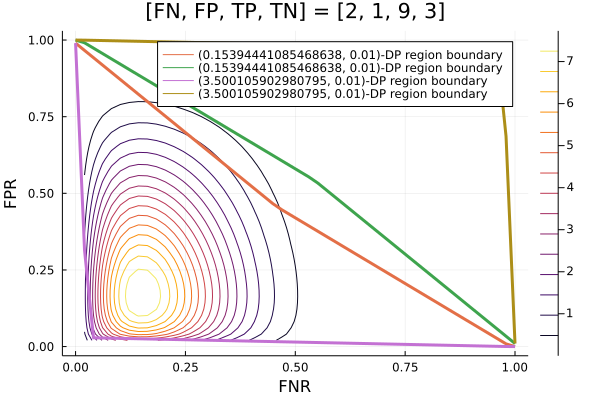}} \\
  \subfloat[$\Pr{\tilde{\mathcal{R}}(0.59, 1.5, 0.01)=0.95}$]{\includegraphics[width=0.48\textwidth]{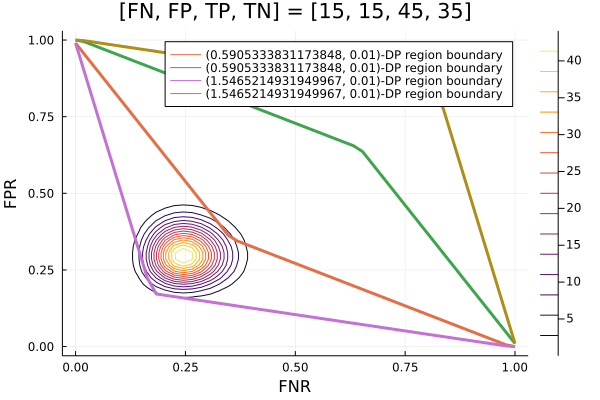}}
  \subfloat[$\Pr{\tilde{\mathcal{R}}(0.45, 0.80, 0.01)=0.95}$]{\includegraphics[width=0.48\textwidth]{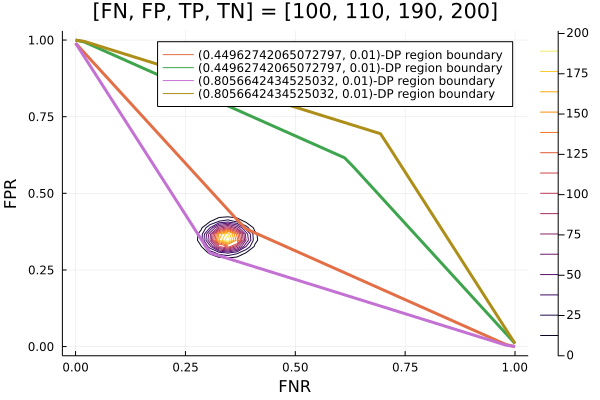}}
  \caption{Convergence of the joint posterior $f_{\FNRFPR}$ as the number of samples grows.}
  \label{fig:example_bayes}
\end{figure}

The results are illustrated in Figure \ref{fig:example_bayes}. 
Initially, we look at privacy regions after only 4 trials.
As expected, the two-sided privacy region is fairly large and covers almost the entire unit square.
As we see more and more samples our confidence increases and the two-sided privacy region shrinks.